\documentclass[twoside,11pt]{article}

\usepackage{jmlr2e}
\usepackage{xcolor}
\usepackage[normalem]{ulem}

\ShortHeadings{Community detection in multilayer networks }{Li et al.}
\firstpageno{1}

\begin{document}

\title{rMultiNet: An R Package For Multilayer Networks Analysis}

\author{\name Ting Li \email tingeric.li@polyu.edu.hk \\
        \addr Department of Applied Mathematics, The Hong Kong Polytechnic University\\
       \name Zhongyuan Lyu \email zlyuab@connect.ust.hk\\
       \addr Department of Mathematics, The Hong Kong University of Science and Technology\\
       \name Chenyu Ren \email chenyu.ren@connect.polyu.hk\\
       \addr Department of Applied Mathematics, The Hong Kong Polytechnic University\\
    \name Dong Xia \email madxia@ust.hk\\
       \addr Department of Mathematics, The Hong Kong University of Science and Technology\\}

\editor{}

\maketitle

\begin{abstract}
This paper develops an R package \textbf{rMultiNet} to analyze multilayer network data. We provide two general frameworks from recent literature, e.g. mixture multilayer stochastic block model(MMSBM) and mixture multilayer latent space model(MMLSM) to generate the multilayer network. We also provide several methods to reveal the embedding of both nodes and layers followed by further data analysis methods, such as clustering. Three real data examples are processed in the package. The source code of \textbf{rMultiNet} is available at
https://github.com/ChenyuzZZ73/rMultiNet.
\end{abstract}


\section{Introduction}
\begin{figure}
  \centering
  \includegraphics[width=6in]{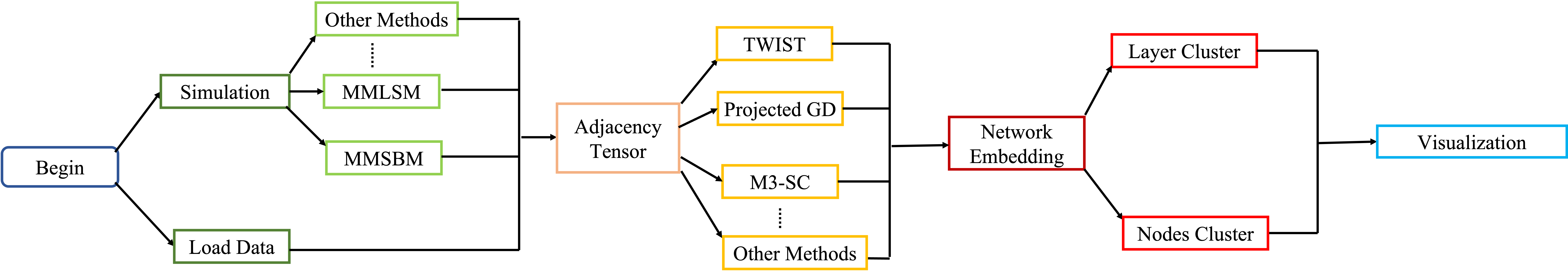}
  \caption{overview of the \textbf{rMultiNet} package} \label{fig:framework}
\end{figure}

The recent decade has witnessed a fast-growing demand for processing and analyzing complex networks. While there are numerous studies about the single static network(\cite{amini2013pseudo, gao2017achieving,gao2018community,wang2021fast,jing2022community,yu2022collaborative}), researchers have shown increasing interest in the study of the multilayer network (\cite{paul2017spectral,le2018estimating,lei2020consistent, arroyo2021inference, jing2021community, li2021super, chen2022global}), which is a more powerful representation of multi-relational data. Numerous kinds of real-world data could be recorded as multilayer networks, such as brain connectivity networks, gene-gene interactivity networks and world trading networks.\\
 Since community structure is a common observation in static network analysis, it naturally raises the question on how to  define and detect community structure in multilayer networks. In a multilayer network, the nodes represent individuals of interest and the edges between nodes in different layers represent different relationships. Such complex relations in multilayer networks pose great challenges to identify and analyze its community structure. Particularly, the heterogeneity across layers can be characterized by individual links, group memberships of nodes, or connectivity patterns inside the community and among different communities. Recently, \cite{jing2021community} proposed a novel mixture multilayer SBM(MMSBM) with a new tensor-based method TWIST to simultaneously cluster networks and identify global and local group memberships of vertices. Moreover, \cite{lyu2021latent} introduced a novel mixture multilayer LSM(MMLSM) that estimates the latent positions of nodes via the generalized low-rank tensor decomposition.\\
In this paper, we propose an R package \textbf{rMultiNet} to analyze the mixture multilayer network. Fig.\ref{fig:framework} illustrates the overview of our package \textbf{rMultiNet}. We provide several prevalent methods for users to study the latent features of multilayer networks. First, we provide three real datasets as examples to study and two generative models, the MMSBM and the MMLSM, to generate mixture multilayer networks. Then, we provide several fitting methods from \cite{jing2021community} and \cite{lyu2021latent} to reveal the embedding results of the multilayer network following with several prevalent clustering methods to analyze the embedding results. Last, several data visualization functions are provided to present the out-comings.    

\section{Package overview}
 \textbf{rMultiNet} includes models to generate multilayer networks, several algorithms to learn complex mixture multilayer networks' latent structures, multiple clustering methods to further analyze the embedding results, and several visualization functions for presentation. The package is organized into the modules listed below:
\begin{itemize}
    \item \textbf{Generation}: \textbf{rMultiNet} adopts the mixture multilayer stochastic(MMSBM) (\cite{jing2021community}) and mixture multilayer latent space(MMLSM) (\cite{lyu2021latent}) to generate mixture multilayer networks.
    \item \textbf{Embedding}: \textbf{rMultiNet} contains two tensor decomposition algorithms for mixture multilayer network embedding, e.g., the TWIST proposed by \cite{jing2021community} and the ProjectedGD introduced by \cite{lyu2021latent}. Moreover, a naive method, spectral clustering on the sum of adjacency matrices from all layers (Sum–Adj) and Mode-3 flatting(M3-SC), proposed in \cite{jing2021community}, is included as a baseline for comparison. 
    \item \textbf{Clustering}:  \textbf{rMultiNet} provides several clustering methods to analyze the embedding results, such as K-means (\cite{likas2003global}), spectral clustering (\cite{dong2012clustering}) and Density-based spatial clustering of applications with noise (DBSCAN) (\cite{hahsler2019dbscan}).
    \item \textbf{Datasets}: \textbf{rMultiNet} contains three datasets for study. 

\textbf{Human malaria parasite gene network}: The data under investigation are the 9 Highly variable regions of the malaria parasite gene sequence. Each network contains 212 nodes, which appear on all 9 layers. More details about the background and the data pre-processing could be found in \cite{larremore2013network} and \cite{jing2021community}.\\
\textbf{Worldwide food trading network}: In this multilayer network, layers represent different products, nodes are countries, and edges at each layer represent trading relationships of a specific food product among countries. The data is collected by \cite{de2015structural} and is available at http://www.fao.org. After data pre-processing (\cite{jing2021community}), we obtain a 30-layers network with 99 nodes. \\
\textbf{UN Commodity trading network}: The dataset contains annual trade information for countries in 2019 from the UN Comtrade Database (https://comtrade.un.org). We focus on the top representative 48 countries ranked by the exports of goods and services in US dollars. Each layer represents a different type of commodity classified into 97 categories (\cite{lyu2021latent}).\\
\end{itemize}

\section{Functionality and Examples}
In this section, we provide the detailed usage of \textbf{rMultiNet} package. Specific meanings of notations used in this part can be found in \cite{jing2021community} and \cite{lyu2021latent}. 
The multilayer network to be explored can be either generated from the package or loaded from external files. The data load from external files needs to be in the form of the `\emph{tensor}` class defined in package \textbf{rTensor}. We provide two approaches to generate the adjacency tensor of the mixture multilayer network in \textbf{rMultiNet} as follows, corresponding to MMSBM and MMLSM respectively.\\
$>$ \emph{library(rMultiNet)}\\
$>$ \emph{GenerateMMSBM(n, m, L, K, d = NULL, r = NULL)}\\
$>$ \emph{GenerateMMLSM(n, m, L, rank, U\_mean= 0.5, cmax =1, d, int\_type = `Uniform', kernel\_fun = `logit', scale\_par=1)}\\
Here, $n$ is the number of vertices, $m$ is the number of types of the network, $L$ is the number of layers, $K$ is the number of groups of vertices, $d$ is the average degree of the network in each layer and 
$r$ is the out-in ratio in each layer. Particularly in  function \emph{GenerateMMLSM}, $rank$ is the rank of latent position matrix $U$, {\emph{U\_mean}} is the mean of the normal distribution of  each entry of $U$, \emph{cmax} is the entry-wise upper bound of core tensor C, {\emph{int\_type}} represents the ways of generating tensor C (\emph{`Uniform'} or \emph{`Norm'}), {\emph{kernel\_fun}} is the link function of generating the adjacency tensor (\emph{`logit'} or \emph{`probit'}) and  {\emph{scale\_par}} is the scaling factor of the parameter tensor. The output is a list including an adjacency tensor and the generating parameters $\Theta$.\\
The multilayer network loaded or generated from the above, will be stored as an adjacency tensor. \textbf{rMultiNet} provides algorithms to learn the latent structure of the adjacency tensor.\\
$>$ \emph{InitializationMMSBM(tnsr, ranks=NULL)}\\
$>$ \emph{PowerIteration(tnsr, ranks=c(2,2,2), type="TWIST", U\_0\_list, delta1=1000, delta2=1000, max\_iter = 25, tol = 1e-05)}\\
The function \emph{InitializationMMSBM} outputs initialization \emph{U\_0\_list}, which can be  the input of function \emph{PowerIteration}. Here \emph{tnsr} is the adjacency tensor of the network,  \emph{type} specifies the iterative algorithm to run \emph{`TWIST'} or \emph{`Tucker'}, \emph{delta1} and \emph{delta2} are tuning parameters for regularization in mode-1 and mode-2, \emph{max\_iter} is the max times of iteration and \emph{tol} is the convergence tolerance. Note that the \emph{ranks} is the rank of the core tensor calculated by the equation $m \times K - (m - 1)$ (see \cite{jing2021community}). The output is a list including the core tensor $Z$, network embedding and node embedding. \\
$>$ \emph{SpecClustering(tnsr, rank, embedding\_type = "Layer")}\\
In function \emph{SpecClustering}, \emph{tnsr} is the adjacency tensor, \emph{rank} is the number of columns of the output matrix $U$, \emph{embedding\_type} specifies the  embedding type (Sum-Adj for \emph{`Node'} and  M3-SC for \emph{`Layer'}). The output matrix $U$ can be applied in cluster methods like kmeans.\\
$>$ \emph{InitializationLSM(gen\_list, n, rank, M, perturb = 0.1, int\_type)}\\
$>$ \emph{ProjectedGD(Ini\_list, Cmax, eta\_outer = 1e-04,
                           tmax\_outer = 35, p\_type =`logit', rd =`Non', show = TRUE, sgma =1, sample\_size =5000)}\\           
In function \emph{InitializationLSM}, \emph{gen\_list} is a list including the adjacency tensor and the parameter $\Theta$ of the mixture multilayer network, \emph{$n$} is the number of nodes, \emph{rank} is the rank of $U$; \emph{M} is the number of network types, \emph{perturb} specifies the upper bound of Uniform distribution, \emph{int\_type} specifies the method to initialize $U$ and $W$ ( \emph{`spec'}, \emph{`rand'} or  \emph{`warm'}). The output of function \emph{InitializationLSM} is a list including the adjacency tensor, $U_{0}$, $W_{0}$ and tuning parameters $\{\delta_{1}, \delta_{2}, \delta_{3}\}$. In function \emph{ProjectedGD}, \emph{Ini\_list} is the output of function \emph{InitializationLSM}, \emph{Cmax} is the upper limits for adding the coefficient constraint, \emph{eta\_outer} is the learning rate in gradient descent, \emph{tmax\_outer} is the number of iterations in gradient descent, \emph{p\_type} specifies the type of link function (\emph{`logit'}, \emph{`probit'} or \emph{`poisson'}), \emph{rd} specifies whether to use stochastic sampling (\emph{`rand'} or \emph{`Non'}) and \emph{sgma} is the link function parameter $\sigma$. The output is the embedding results of nodes and layers.\\
\textbf{rMultiNet} also implements functions for visualization.\\
$>$ \emph{Embedding\_network(network\_membership,L, paxis = 2)} \\
This function is used to produce plots of network embedding and node embedding, respectively. Here \emph{paxis} specifies the number of eigenvectors to use in the plot. If the number of eigenvectors is more than two, a plot table is supposed to generate. By default, it plots the second eigenvector and the third eigenvector.\\
$>$ \emph{Community\_cluster\_km(embedding,type,cluster\_number)} \\
$>$ \emph{Community\_cluster\_dbscan(embedding,type,eps\_value =.05,pts\_value=5)} \\
Clustering algorithms like K-means and DBSCAN can be applied to the results of embedding. Here, \emph{type} can  be either node embedding \emph{`n'} or network embedding \emph{`N'}, \emph{cluster\_number} is the number of clusters for Kmeans, \emph{eps\_value} and \emph{pts\_value} is  parameters for DBSCAN.

\section{Summary}
\textbf{rMultiNet} introduces an extension R package that includes a variety of traditional and state-of-the-art tensor decomposition methods for mixture multilayer network analysis. The package is developed with the modular pipeline mode: generative modeling, embedding algorithms, and visualization. In response to growing data and interactions in different networks, \textbf{rMultiNet} aims to help study complex networks, especially mixture multilayer networks. The dynamic networks also apply in this package which can be regarded as a special case of multilayer networks with layers indexed by time.


\acks{}


\vskip 0.2in

\bibliography{sample}

\end{document}